\documentclass[runningheads]{llncs}

\usepackage{eccv}

\usepackage{eccvabbrv}

\usepackage{graphicx}
\usepackage{booktabs}
\usepackage{comment}
\usepackage{lscape} 
\usepackage{amssymb}
\usepackage{pifont}
\newcommand\mypar[1]{\par\noindent\textbf{#1}\;\;}

\usepackage[accsupp]{axessibility}  

\usepackage{hyperref}

\usepackage{orcidlink}

\begin{document}

\title{JDT3D: Addressing the Gaps in LiDAR-Based Tracking-by-Attention}

\titlerunning{JDT3D}

\author{Brian Cheong$^*$\orcidlink{0009-0000-4110-6504} 
\and
Jiachen Zhou$^*$\orcidlink{0009-0006-1234-1883} 
\and
Steven Waslander\orcidlink{0000-0003-4217-4415}
}
\def\thefootnote{*}\footnotetext{Denotes equal contribution.}\def\thefootnote{\arabic{footnote}}

\authorrunning{B.~Cheong et al.}

\institute{University of Toronto\\\email{\{brian.cheong, jason.zhou, steven.waslander\}@robotics.utias.utoronto.ca}}


\maketitle

\begin{abstract}
Tracking-by-detection (TBD) methods achieve state-of-the-art performance on 3D tracking benchmarks for autonomous driving. On the other hand, tracking-by-attention (TBA) methods have the potential to outperform TBD methods, particularly for long occlusions and challenging detection settings. 
This work investigates why TBA methods continue to lag in performance behind TBD methods using a LiDAR-based joint detector and tracker called JDT3D. Based on this analysis, we propose two generalizable methods to bridge the gap between TBD and TBA methods: track sampling augmentation and confidence-based query propagation.
JDT3D is trained and evaluated on the nuScenes dataset, achieving 0.574 on the AMOTA metric on the nuScenes test set, outperforming all existing LiDAR-based TBA approaches by over 6\%. 
Based on our results, we further discuss some potential challenges with the existing TBA model formulation to explain the continued gap in performance with TBD methods.
The implementation of JDT3D can be found at the following link: \href{https://github.com/TRAILab/JDT3D}{https://github.com/TRAILab/JDT3D}.
\end{abstract}

\section{Introduction}
\label{sec:intro}

Multi-object tracking (MOT) is an essential component in the perception system of autonomous vehicles, as it allows autonomous agents to reason and plan appropriately in dynamic environments. The task involves identifying object trajectories, requiring that each object is accurately detected and consistently identified over time within the scene.

The majority of MOT methods adhere to the tracking-by-detection (TBD) paradigm. This paradigm is comprised of two separate processes: generating bounding box predictions without track IDs and associating these predictions to a set of maintained tracks based on some matching criteria.
TBD methods currently achieve state-of-the-art results on 3D MOT benchmarks \cite{Li2023PolyMOTAP, Wang2022CAMOMOTCA, Liu2022BEVFusionMM}. 

In contrast to the decoupled approach of TBD, joint detection and tracking (JDT) approaches have attempted to unify the detection and tracking tasks in an end-to-end manner \cite{Pang_2023_CVPR, Zhang2023MotionTrackET, Zhang2022MOTRv2BE, trackformer_2022_CVPR}. 
Recent works in autonomous driving perception \cite{hu2023planning, Mahmoud_2023_WACV} and computer vision~\cite{huang2023video} demonstrate that networks trained end-to-end have shown superior performance compared to pipeline-based techniques, despite initially lower performance from earlier works \cite{Zeng2021MOTREM, Sun2020TransTrack}. End-to-end methods can learn richer and more generalizable representations while minimizing the task-specific engineering of individual modules required in a pipeline-based approach. While TBD methods enforce a unidirectional flow of information from the detector to the tracker, the JDT formulation allows both the detector and tracker to exchange useful information to enhance the performance of both tasks.

\begin{figure}[tb]
    \centering
    \includegraphics[width=0.75\linewidth]{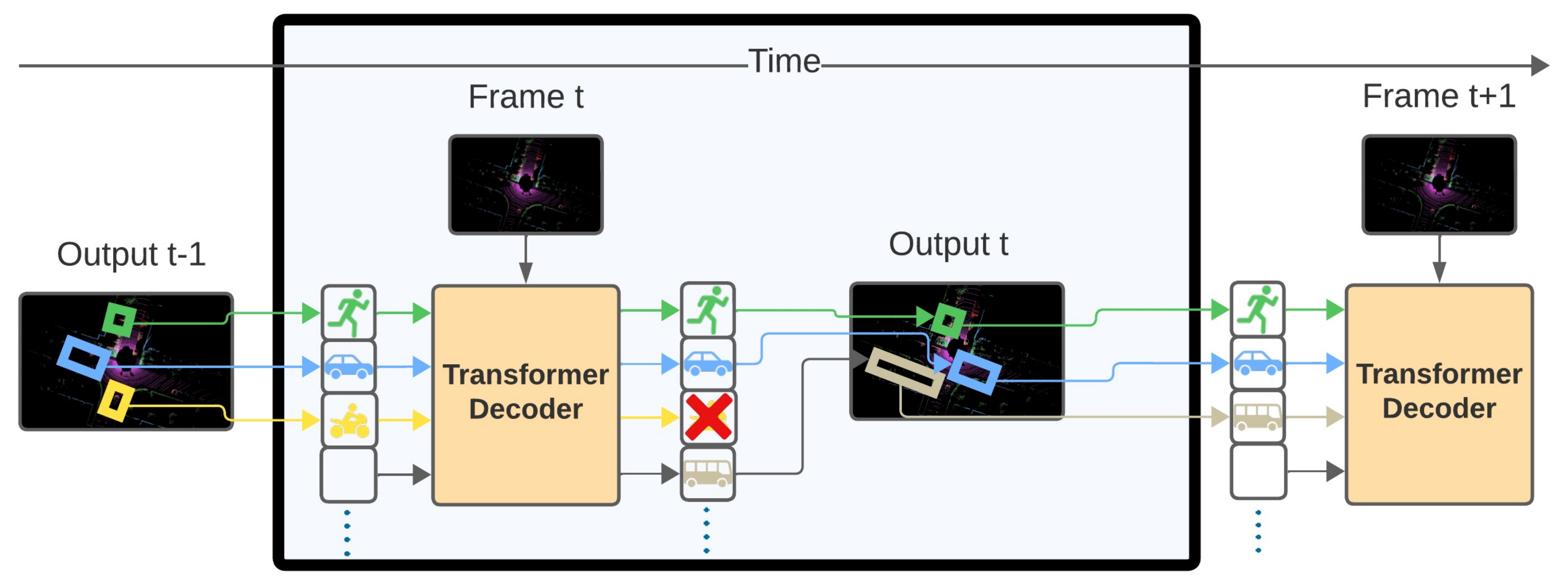}
    \caption{Illustration of tracking by attention. Each object is represented by a query. In the diagram, the yellow motorcycle leaves the frame, so it is removed from the set of maintained queries. To detect new objects in the scene, such as the tan bus, proposal queries are appended at each time step, represented by the blank squares.}
    \label{fig:TBA_simplified}
\end{figure}

Integrating detection and tracking into a joint learning task requires a unique approach to formulating the problem and training the model. One such approach is tracking-by-attention (TBA), illustrated in \cref{fig:TBA_simplified}. In TBA, objects are represented as vector embeddings, or "queries", that are used to detect the same object over multiple frames.
While recent TBA methods \cite{Pang_2023_CVPR, Zhang2023MotionTrackET, Zhang2022MOTRv2BE, trackformer_2022_CVPR} have been proposed as alternatives to TBD, they continue to underperform on the MOT task, especially in the LiDAR domain \cite{Zhang2023MotionTrackET}. However, they have shown potential in certain cases, such as significantly fewer ID-switch errors \cite{Pang_2023_CVPR}. The unexplored potential of TBA in the LiDAR domain motivates our work in addressing a key question: what is holding back LiDAR-based TBA?
In this work, we propose a LiDAR-based TBA tracking method called JDT3D, based on which we explore different aspects that hold back LiDAR-based TBA.

One insight is that LiDAR-based tracking methods can suffer from sparse supervision signals due to point cloud and object sparsity. Inspired by object sampling augmentation \cite{Yan2018SECONDSE} for LiDAR-based detection methods, we propose track sampling, a temporally consistent augmentation method that injects consistent objects over multiple LiDAR frames to enrich supervision signals while maintaining temporal consistency.

In addition, existing TBA methods maintain an inconsistency between training and inference regarding their query propagation method. During training, queries that are matched to ground truth samples are propagated to the next frame, but at inference, propagation is based on a confidence threshold. This inconsistency between training and inference could confuse the model to over-trust false positive queries. To address this, we propose a confidence threshold propagation strategy consistent for both training and inference and conduct comprehensive evaluations and discussions on the strategy design.

We summarize our contributions as follows:

\mypar{JDT3D} We present JDT3D, a LiDAR-based tracking-by-attention model that outperforms existing LiDAR-based tracking-by-attention methods on the nuScenes test set by over 6\% on the AMOTA metric.

\mypar{Confidence-based query propagation} We perform ablations to validate the effectiveness and generalizability of the confidence-based propagation criterion.

\mypar{Track Sampling Augmentation} We propose and validate a novel data augmentation technique for multi-frame LiDAR methods that significantly improves performance and convergence rates during training.

\mypar{Analysis} Based on our experiments using JDT3D, we provide a detailed analysis to explain the continued gap in performance between LiDAR-based TBD and TBA.
We attribute the gap to the reduced detection performance when training on multiple consecutive frames and the handling of queries that results in "temporal confusion".
\section{Related Works}
\label{sec:related_works}

Multi-object tracking (MOT) methods either follow the tracking-by-detection paradigm, in which object detection is independently performed on single frames before the tracking module, or they perform detection and tracking jointly in an end-to-end fashion. 
MOT can also be further categorized into 2D and 3D MOT. The task of 2D MOT involves predicting object tracks in 2D space, while 3D MOT predicts the tracks in 3D space. 2D MOT methods generally take in camera images as sensor input, while 3D MOT methods may take in camera images, LiDAR point clouds, or both. In this work, we will be focusing our analysis on 3D MOT methods.

\subsection{Tracking-by-detection (TBD)}
In the tracking-by-detection paradigm, the detection and tracking problems are solved by two distinct modules. First, a single-frame object detector is used to detect objects in a given scene. Next, the detections are associated with a set of maintained tracks by a tracking module and used to update them with current detection information. Unassigned detections are used to initialize new tracks.

While different tracking methods exist among TBD methods, they are all consistently tied to the detector's performance. Several top-performing 3D tracking methods \cite{Bai_2022_CVPR, Liu2022BEVFusionMM, Chen2023FocalFormer3DF} are strong detectors that have been paired with the simple tracking method of CenterPoint \cite{Yin_2021_CVPR}. The formulation of TBD inhibits the detector from taking advantage of temporal object information that would enhance object detection. As well, errors in the detection are guaranteed to be propagated downstream to the tracking task. This motivates the exploration of methods that can jointly optimize detection and tracking to achieve MOT.

\subsection{Joint Detection and Tracking (JDT)}
Joint detection and tracking (JDT) is a recent paradigm in both 2D and 3D MOT that involves training a model to perform detection and tracking end-to-end. In contrast to the TBD methods that rely on a dedicated detector to generate detections, JDT methods simultaneously output object detections with associated tracking IDs. Without heuristic matching or hand-crafted rules, SimTrack \cite{Luo_2021_ICCV} builds upon CenterPoint \cite{Yin_2021_CVPR} by learning a hybrid-time centerness map between two consecutive LiDAR point clouds. AlphaTrack \cite{iros2021_alphatrack} adds a cross-modal fusion scheme to the CenterPoint tracking method\cite{Yin_2021_CVPR} and applies an alternating training strategy to balance between the detection branch and appearance affinity. The above two methods perform detection and tracking jointly, but known tracks are not utilized to inform future detections. 3D DetecTrack \cite{Koh_Kim_Yoo_Kim_Kum_Choi_2022} proposes tracklet-aware 3D detection that reconfigures the initial detection using the latest tracklets' information and employs a spatio-temporal gated GNN for association.

\subsubsection{Tracking-by-Attention (TBA)}
A growing trend within JDT is the tracking-by-attention (TBA) approach that emerged from TrackFormer \cite{trackformer_2022_CVPR} for 2D MOT. TBA uses the transformer architecture \cite{NIPS2017_attention} and formulates tracking as a frame-to-frame set prediction problem that can be trained in an end-to-end manner.

TrackFormer \cite{trackformer_2022_CVPR} extends a 2D query-based detector, DETR \cite{carion2020detr} and its variant Deformable DETR \cite{deformabledetr}, by designing autoregressive track queries that are passed across frames with associated object identities. MOTRv2 \cite{Zhang2022MOTRv2BE} uses a pretrained YOLOX detector to initialize their object queries and uses the detections to anchor the decoded bounding box positions. In addition, track queries during training are propagated based on the prediction confidence rather than whether a prediction was matched to a ground truth object. This naturally creates false positive and false negative cases during training, removing the need for query injection and removal augmentations \cite{trackformer_2022_CVPR, Zeng2021MOTREM}. 
JDT3D applies confidence-based track query propagation to LiDAR-based TBA, and we perform ablations showing the method's improved performance over the ground truth matching criterion.

MUTR3D \cite{Zhang_2022_CVPRW} and PF-Track \cite{Pang_2023_CVPR} demonstrate the transfer of the tracking-by-attention framework to multi-view vision-only 3D MOT.
PF-Track \cite{Pang_2023_CVPR} also designs past and future reasoning modules to enhance the spatio-temporal representation of queries. MotionTrack \cite{Zhang2023MotionTrackET} investigates the feasibility of extending the TBA paradigm to LiDAR-based and LiDAR-camera fusion-based 3D JDT by building upon TransFusion \cite{Bai_2022_CVPR}, a query-based 3D object detector. Rather than implicitly learning which queries should be suppressed through the transformer decoder, MotionTrack learns an explicit affinity matrix between tracks and objects in the scene and performs greedy matching to determine their association. This explicit matching echoes the decoupling found in TBD methods and potentially detracts from the benefits of an end-to-end method. While vision-only TBA methods have shown surprising results on the MOT task, the baseline results set by MotionTrack's LiDAR-only and fusion TBA methods underperform compared to LiDAR-based TBD baselines such as CenterPoint \cite{Yin_2021_CVPR} and SimpleTrack \cite{pangSimpleTrackUnderstandingRethinking2023}, prompting further exploration of LiDAR TBA methods. Our work explores different potential hypotheses for the reported performance gap between LiDAR-based TBD and TBA methods and proposes improvements to the existing TBA methods.

\subsection{LiDAR Data Augmentations}
As with other machine learning tasks, data augmentation can greatly enhance the performance of LiDAR-based 3D object detection methods. Some common augmentations include random flipping, rotation, translation, and scaling of the point cloud \cite{Hahner_2020_CoRR}. In addition to these, object sampling is also a widely used augmentation method for LiDAR-based object detectors \cite{Bai_2022_CVPR, Liu2022BEVFusionMM, Yin_2021_CVPR}. First introduced in SECOND \cite{Yan2018SECONDSE}, object sampling involves creating a database of ground truth object point cloud scans and bounding boxes and randomly sampling and injecting them into the scene during training to improve the convergence and performance overall. However, this augmentation does not apply directly to training TBA methods, since it would result in sporadic track instances in the scene in each training sequence.
Our proposed track sampling augmentation samples consecutive track instances and injects them into a sequence of frames during training. This maintains the temporal nature of the object tracks while still offering the same benefits of faster convergence.
\section{Method}
\label{sec:method}

\begin{figure*}[tb]
    \includegraphics[width=0.9\textwidth]{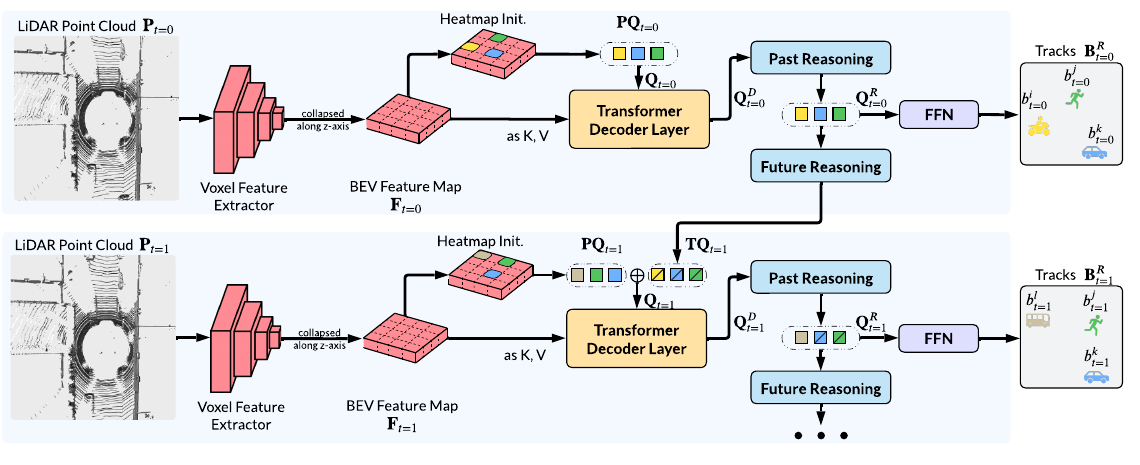}
    \centering
    \caption{JDT3D Architecture. At each time step, a BEV feature map is extracted and used to initialize a set of proposal queries. The proposal queries are concatenated to track queries passed from the previous frame and used to predict objects in the scene. Track queries detect the same unique objects in each time step, while proposal queries detect untracked or new objects.}
    \label{fig:sys_arch}
\end{figure*}

To perform our analysis, we developed our own baseline LiDAR-based TBA method since no such implementations were made available as of the writing of this paper. We build on aspects from query-based 3D object detection \cite{Bai_2022_CVPR} and vision-only TBA \cite{Pang_2023_CVPR} to formulate our method, called JDT3D. JDT3D is summarized in \cref{fig:sys_arch}.

\subsection{Overview}
At each time step, a feature extractor takes in the LiDAR scan of the scene, $\textbf{P}$, and converts it into a BEV feature map representation, $\textbf{F}$, using a standard voxel feature extractor \cite{Zhou2017VoxelNetEL}. This feature map is used to predict heatmaps of potential object locations and initialize a set of proposal queries, $\textbf{PQ}$ \cite{Bai_2022_CVPR}.

In the first time step, the proposal queries are passed through a transformer decoder layer with the BEV feature map to obtain a set of updated queries, $\textbf{Q}^D$. The past reasoning module refines the updated queries with historical track information \cite{Pang_2023_CVPR} before a feedforward network decodes the refined queries, $\textbf{Q}^R$ into bounding boxes and class predictions, $\textbf{B}^R$. During training, $\textbf{Q}^D$ is decoded into a set of bounding boxes and class predictions, $\textbf{B}^D$, for an auxiliary loss.

To create the track queries for the next frame, the predictions and their corresponding queries are filtered with a confidence threshold. The filtered queries are passed to the future reasoning module \cite{Pang_2023_CVPR} which predicts a motion trajectory and projects each query into the next time step. 

At every following time step, new proposal queries are initialized with the heatmap as described above and are concatenated with track queries from the previous time step, $\textbf{TQ}$, before being passed to the transformer decoder. 

To perform tracking, high-confidence predictions and associated queries are assigned an ID. Track queries are used to detect the same object consistently over multiple frames, while proposal queries are used to detect objects to be added to the track query set. Tracking is achieved implicitly through the direct relationship between each prediction and each query passed into the transformer decoder. Trajectories can be obtained by tracing a given track query back in time.

\subsection{Query Propagation}
\label{sec:query_prop}

During training, only the refined queries corresponding to predictions with a confidence score greater than $\tau_{pass}$ are passed to the next time step. This method of passing and pruning queries has the advantage of naturally creating instances of false negative and false positive track queries during training \cite{Zhang2022MOTRv2BE}. 
False negative track queries are missed tracks that should be re-detected in the next frame as if they were new objects entering the scene. 
False positive track queries simulate tracks that have left the scene and should be pruned in the next time step. 

This method has the benefit of better matching the inference query propagation behaviour. As well, it creates more instances during training where the decoder must suppress track query predictions and output positive proposal query predictions, ensuring robust handling of track births and deaths.

\subsection{Ground Truth Assignment}
The ground truth assignment of the decoder predictions, $\mathbf{B}_t^D$, and the refined predictions, $\mathbf{B}_t^R$, are both done in the same manner using a two-stage assignment process. The first stage involves matching the track queries based on their previously matched ground truth objects \cite{trackformer_2022_CVPR}. If the previously matched ground truth track is not present in the current time step, such as in the case of a track leaving the scene, then the track query prediction is assigned to no object. False positive track queries were never assigned a ground truth track previously, so they are also assigned to no object.

The second stage of the assignment involves matching the proposal queries to the unmatched ground truths. The unmatched ground truths would include new objects in the scene and missed objects from previous time steps. A Hungarian matcher is used to compute the optimal assignment between the unmatched queries and ground truths based on an association cost \cite{Kuhn1955TheHM}.

\subsection{Track Sampling Augmentation}
\label{subsec:track_sampling}
\begin{figure}[tb]
    \centering
    \includegraphics[width=0.65\linewidth]{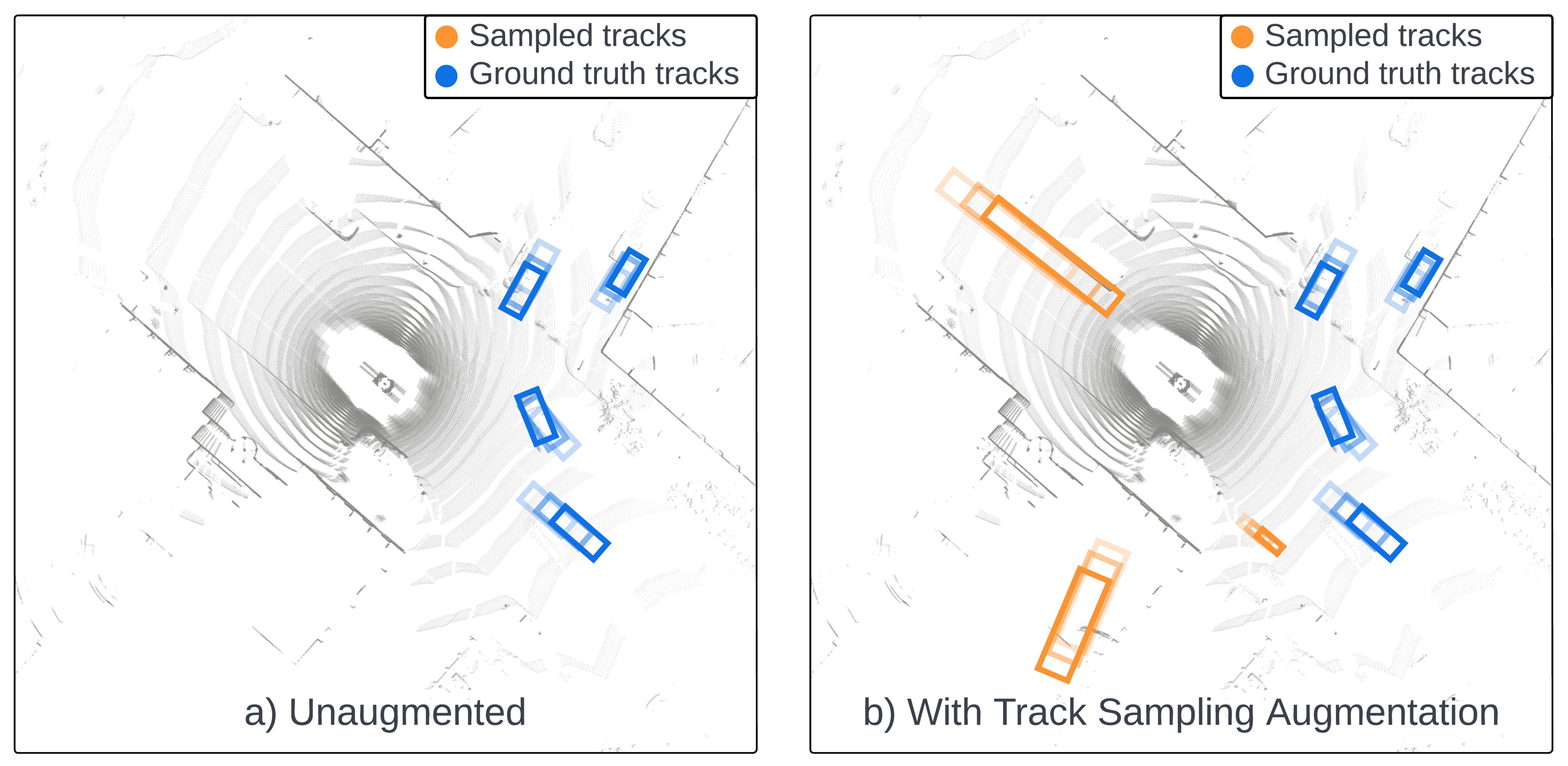}
    \caption{An example of track sampling augmentation over three consecutive frames from the nuScenes dataset. The original trajectories and sampled trajectories are shown in blue and orange, respectively.
    Only a subset of the tracks and the third LiDAR scan are shown for visual clarity. The older boxes are shown with transparency.}
    \label{fig:track-sampling-vis}
\end{figure}

To improve the convergence rate and performance, we introduce track sampling augmentation during training. 
Inspired by the object sampling augmentation \cite{Yan2018SECONDSE}, our track sampling involves adding additional tracks into a training clip from a database of tracks in a temporally consistent manner, as illustrated in \cref{fig:track-sampling-vis}. The original object sampling augmentation does not account for temporal consistency, and applying it naively to our multi-frame training scheme would result in disjointed track instances during training.

First, the training set is preprocessed to generate a database of ground truth tracks, including bounding boxes, object classes, track IDs, and LiDAR points contained in the bounding boxes.
To perform track sample augmentation on a training clip, a set of tracks is randomly sampled, avoiding track IDs that overlap with the existing ground truth objects in the scene. This enforces unique IDs for the tracks in the sequence. 
To inject a track, $L$ consecutive ground-truth boxes are randomly selected from each sampled track. In each training frame, the LiDAR points corresponding to the sampled objects are used to replace regions in the original point cloud. The objects are injected in chronological order over the $L$ frames. A collision check is performed to ensure that there are no collisions between injected objects and the original ground truth objects in the scene. In the case of a collision, the would-be sampled object is pruned from that frame, simulating an occlusion.

The benefits of track sampling mirror those of object sampling \cite{Yan2018SECONDSE}, with the extra ground truth objects in the scene providing more samples to learn from in a single frame. This results in a stronger training signal and more varied scenes.

\subsection{Network Training and Losses}
To train JDT3D, $L$ LiDAR frames are passed sequentially to the model as a training sample. 
The loss for a given frame, $\mathcal{L}_i$, is computed as the weighted sum of the following losses:
\begin{equation}
    \mathcal{L}_i = \lambda_{h} \mathcal{L}_{h} + \lambda_f \mathcal{L}_f + \sum_{m=\{D, R\}} (\lambda_{reg}^m \mathcal{L}_{reg}^m + \lambda_{cls}^m \mathcal{L}_{cls}^m)    
\end{equation}
The superscripts $D$ and $R$ correspond to predictions output by $\textbf{Q}^D$ and $\textbf{Q}^R$, respectively. $\mathcal{L}_{reg}^m$ is the L1 loss over the bounding boxes, and $\mathcal{L}_{cls}^m$ is the focal loss \cite{Lin2017FocalLF} over the classification predictions. $\mathcal{L}_{h}$ is the Gaussian Focal Loss on the predicted object heatmaps \cite{CornerNet, Yin_2021_CVPR}, and $\mathcal{L}_f$ is the L1 loss over the motion trajectory predicted by the future reasoning module \cite{Pang_2023_CVPR}. Each loss has an associated scalar weight term, $\lambda$.
The total loss is computed as the sum of each indivudal frame loss:
\begin{equation}
    \mathcal{L}_{clip} = \sum_{i=1}^L \mathcal{L}_i
\end{equation}
\section{Experiments}
\label{sec:Experiments}

\subsection{Dataset and Metrics}
JDT3D is trained and evaluated on the nuScenes dataset \cite{Caesar2019nuScenesAM}, a multi-modal autonomous driving dataset containing data from six multi-view cameras, five radars, and a 360-degree LiDAR. The dataset contains a total of 40,157 frames across 1000 scenes, of which 850 scenes are annotated for training (700) and validation (150) at a frame rate of 2 Hz. We train on the seven object classes annotated for the tracking task.

To evaluate JDT3D, the average multi-object tracking accuracy (AMOTA) and average multi-object tracking precision (AMOTP) tracking metrics \cite{Weng2020_AB3DMOT} are computed across the seven tracking classes of nuScenes, evaluating the tracking accuracy and tracking position misalignment, respectively. Secondary metrics are also computed by the nuScenes MOT benchmark \cite{Caesar2019nuScenesAM}, such as the number of false positives, false negatives, and identity switches. We also evaluate the detection performance across the tracking classes, with the mean average precision (mAP) and nuScenes Detection Score (NDS) metrics.

\subsection{Implementation Details}
\label{sec:implementation_details}
VoxelNet \cite{Yan2018SECONDSE, Zhou2017VoxelNetEL} is used as the LiDAR backbone to extract the BEV feature map. For each frame, the top 200 predictions from the heatmap are used to initialize the proposal queries. The training clip length $L$ is set to 3 frames. The confidence threshold for track queries $\tau_{pass}$ is set to 0.4 for training and inference. The Hungarian matching cost for ground truth assignment consists of the sum of the L1 cost on the predicted box and the focal loss \cite{Lin2017FocalLF} on the predicted class confidence. All detections are generated in the car ego-frame, so the object centers of the track queries are updated to their relative positions in the current training frame based on the car's ego-motion between time steps. A pretrained VoxelNet backbone from BEVFusion \cite{Liu2022BEVFusionMM} is used to initialize the feature extractor. To train JDT3D, the model is first trained on the detection task using $L=1$ for 20 epochs. It is then further trained on the tracking task using $L=3$ for another 20 epochs. All experiments were trained and evaluated on eight NVIDIA V100 GPUs using random flipping, rotation, translation, and scaling on the training sequence, and track sampling augmentation, unless otherwise specified. JDT3D performs inference at 3.85 Hz on a Tesla V100.

\subsection{Main Results}

\begin{table}[t]
\centering
\caption{A comparison of 3D multi-object tracking methods evaluated on the nuScenes \cite{Caesar2019nuScenesAM} test set. \textbf{Bold} represents the best results among the Transformer-based TBA methods, while \underline{Underlined} numbers are the best overall.}
\begin{tabular}{c|c|c|ccccc}
\specialrule{1pt}{0pt}{0pt}
Method & TBA & Modality             & AMOTA $\uparrow$  & AMOTP $\downarrow$    & FP $\downarrow$   & FN $\downarrow$ & IDS $\downarrow$ \\ 
\hline
AB3DMOT \cite{Weng2020_AB3DMOT}     & & L & 0.151             & 1.501                 & 15088             & 75730 & 9027         \\
Chui et al. \cite{chiu2020probabilistic}       & & L & 0.550   & 0.798 & 17533 & 33216 & 950 \\
CenterPoint \cite{Yin_2021_CVPR}    & & L & 0.638             & 0.555                 & 18612             & \underline{22928} & 760          \\
SimpleTrack \cite{pangSimpleTrackUnderstandingRethinking2023} & & L & \underline{0.668} & \underline{0.550} & 17514 & 23451 & 575          \\ 
\hline
PF-Track \cite{Pang_2023_CVPR} & \checkmark & C & 0.434 & 1.252 & 19048 & 42758 & \underline{\textbf{249}} \\ 
DQTrack \cite{li2023end} & \checkmark & C & 0.523 & 1.096 & - & - & 1204 \\
MotionTrack-L \cite{Zhang2023MotionTrackET}    & \checkmark & L & 0.51           & 0.99      & - & - & 9705         \\
MotionTrack-LC \cite{Zhang2023MotionTrackET}    & \checkmark & L+C & 0.55           & 0.871      & - & - & 8716         \\
JDT3D (ours)                        & \checkmark & L & \textbf{0.574}             & \textbf{0.837}                 & \underline{\textbf{11152}} & \textbf{29919} & 254 \\
\bottomrule
\end{tabular}
\label{tab:tbd_jdt_nus_test_comparison}
\end{table}

Our main results compare JDT3D with baseline TBD methods and relevant TBA methods evaluated on the nuScenes test set \cite{Caesar2019nuScenesAM}, summarized in \cref{tab:tbd_jdt_nus_test_comparison}.
JDT3D outperforms the current state-of-the-art LiDAR-based TBA methods, with a relative improvement of 12.5\% on the AMOTA metric and 15.5\% on the AMOTP metric on the test split. Moreover, with LiDAR input only, JDT3D also outperforms the fusion-based MotionTrack-LC \cite{Zhang2023MotionTrackET} with a relative improvement of 4.36\% on AMOTA and 3.79\% on AMOTP on the test set.
While JDT3D underperforms compared to baseline TBD methods on most metrics, it achieves the lowest number of ID switches among LiDAR-only methods, indicating a strong track coherence.

JDT3D has significantly more false negatives compared to TBD methods, but fewer false positives. The potential sources of false negatives are either due to missed detections or tracks diverging from the ground truths. However, the latter case would result in both high false negatives and false positives. Based on the low false positive results from JDT3D, we conclude that the false negatives are mainly due to missed detections.


\begin{table}[t]
\centering
\caption{Ablation on the track sampling augmentation, reported on the nuScenes validation set. The mAP metric is computed on the seven tracking categories.}
\begin{tabular}{c|cccc}
\specialrule{1pt}{0pt}{0pt}
    Track Sampling Rate & AMOTA $\uparrow$ & AMOTP $\downarrow$ & IDS $\downarrow$ & mAP $\uparrow$ \\
    \hline
    0.0 & 0.314 & 1.377 & 388 & 0.225 \\
    0.25 & 0.600 & 0.864 & 210 & 0.3867 \\
    0.50 & \textbf{0.622} & 0.816 & 203 & \textbf{0.400} \\
    1.0 & 0.621 & \textbf{0.790} & \textbf{150} & 0.399\\
    \bottomrule
\end{tabular}
\label{tab:sampling_ablation}
\end{table}

\subsection{Ablation Studies}
\label{sec:ablations}

\subsubsection{Track Sampling.}
By comparing the performance of JDT3D on the nuScenes validation set with and without track sampling augmentation, we verify that sampling extra tracks into a training sequence significantly improves the convergence rate and the final performance. The comparison summarized in \cref{tab:sampling_ablation} shows a 97\% relative improvement on the AMOTA metric from track sampling augmentation compared to no track sampling, with a similar trend across all other metrics, including detection metrics such as the mAP. We also analyzed the effect of varying the number of objects that are sampled by reducing the sampling frequency to 0.5 and 0.25 of the nominal rate. We observe that the AMOTA and mAP metrics plateau after a sampling rate of 0.5, although the AMOTP and IDS metrics see continued improvement with increased track sampling. The benefits of implementing track sampling are expected to generalize to other LiDAR-based TBA methods, but there are currently no available implementations to verify this hypothesis.

\subsubsection{Training Clip Length.}
To validate our choice of clip length, the performances of JDT3D models trained with two to four frame sequences were compared and summarized in \cref{tab:clip_len_ablation}. The best performance is achieved when trained over $L=3$ frames, with a significant jump when moving from $L=2$ to $L=3$. We hypothesize that the training process benefits from the longer-term, temporally consistent information, resulting in improved tracking performance. Moving from $L=3$ to $L=4$, the performance drops slightly while also significantly increasing the train time and required memory.

\begin{table}[t]
\centering
\caption{Ablation on the training sequence length, $L$, reported on the nuScenes validation set.}
    \begin{tabular}{c|cccc}
        \specialrule{1pt}{0pt}{0pt}
        Clip Length ($L$) & AMOTA $\uparrow$ & AMOTP $\downarrow$ & IDS $\downarrow$ & mAP $\uparrow$ \\
        \hline
        2 & 0.397 & 1.294 & 446 & 0.214 \\
        3 &  \textbf{0.621} & \textbf{0.790} & \textbf{150} &\textbf{0.399} \\
        4  & 0.599 & 0.783 & 173 & 0.392 \\
        \bottomrule
    \end{tabular}
\label{tab:clip_len_ablation}
\end{table}

\subsubsection{Query Propagation Method.}
In \cref{tab:query_prop_ablation}, we compare different query propagation methods; the first method is passing queries that have been matched to a ground truth \cite{Pang_2023_CVPR, trackformer_2022_CVPR}, and the second is passing based on the $\tau_{pass}$ threshold for high confidence predictions as described in \cref{sec:query_prop}.
Results show that using confidence as a filtering criterion during training enhances the overall tracking metrics AMOTA and AMOTP by a large margin and significantly reduces the number of false positives and false negatives while increasing the number of true positives. While the ID switch errors are higher, this increase is less than $2\%$ relative to the increase in true positives. Thus, we verify that using confidence as the query propagation criterion is more appropriate for the TBA paradigm.

\begin{table}[t]
\centering
\caption{Ablation on the query propagation criteria, reported on the nuScenes \cite{Caesar2019nuScenesAM} validation set. Passing queries with high confidence encourages the re-detection of poor detections and handling of false positives and false negatives.}
    \begin{tabular}{c|cccc}
        \specialrule{1pt}{0pt}{0pt}
        Query Propagation & AMOTA $\uparrow$ & AMOTP $\downarrow$ & Recall $\uparrow$ & IDS $\downarrow$\\
        \hline
        Matched to a GT & 0.464 & 0.917 & 0.511 & \textbf{82} \\
        High Confidence & \textbf{0.621} & \textbf{0.790} & \textbf{0.669} & 150 \\
        \bottomrule
    \end{tabular}
\label{tab:query_prop_ablation}
\end{table}

In addition, we show the generalizability of this query propagation method by performing the same ablation on PF-Track. The results in \cref{tab:query_prop_pftrack_ablation} show improvements across most metrics, with a similar behaviour of more IDS errors.

\begin{table}[tb]
\centering
\caption{An ablation comparing the original ground truth-based propagation and our confidence-based propagation on PF-Track-S \cite{Pang_2023_CVPR} evaluated on the nuScenes \cite{Caesar2019nuScenesAM} validation set.}
    \begin{tabular}{c|cccc}
        \specialrule{1pt}{0pt}{0pt}
        Query Propagation & AMOTA $\uparrow$ & AMOTP $\downarrow$ & Recall $\uparrow$ & IDS $\downarrow$\\
        \hline
        Matched to a GT & 0.408 & 1.343 & 0.507 & \textbf{166} \\
        High Confidence & \textbf{0.426} & \textbf{1.316} & \textbf{0.558} & 306\\ 
        \bottomrule
    \end{tabular}

\label{tab:query_prop_pftrack_ablation}
\end{table}

\subsubsection{Comparison with TBD Baseline.}
We compare the performance between our TBA framework and baseline TBD methods such as the CenterPoint greedy-based tracking method \cite{Yin_2021_CVPR} and SimpleTrack \cite{pangSimpleTrackUnderstandingRethinking2023}.
To compare our tracking method to the TBD trackers independent of detection performance, we pass the detections generated by JDT3D to the tracker and evaluate its tracking performance compared to our joint approach. The results in \cref{tab:jdt3d_tbd_comparison} demonstrate clear advantages of our joint end-to-end approach that offers temporal consistency and data association and reaffirm that the gap in tracking performance lies in the missed detections.

\begin{table}[t]
\centering
\caption{Comparison with "Tracking by Detection" on the nuScenes validation set, using the detections generated by JDT3D.}
    \begin{tabular}{c|ccc}
        \specialrule{1pt}{0pt}{0pt}
        Tracking Method & AMOTA $\uparrow$ & AMOTP $\downarrow$ & IDS $\downarrow$ \\
        \hline
        CenterPoint \cite{Yin_2021_CVPR} & 0.608 & 0.825 & 491 \\
        SimpleTrack \cite{pangSimpleTrackUnderstandingRethinking2023}  & 0.573                & 0.877                & 5585                 \\ \hline
        JDT3D (Ours) & \textbf{0.621} & \textbf{0.790} & \textbf{150} \\
        \bottomrule
    \end{tabular} 
\label{tab:jdt3d_tbd_comparison}
\end{table}

\subsection{Detection Performance}
As mentioned in \cref{sec:implementation_details}, JDT3D is pre-trained to perform single-frame detection and fine-tuned to perform tracking on sequences of multiple LiDAR frames. While the pre-training step achieves reasonable detection performance, the performance drops significantly after fine-tuning, as seen in \cref{tab:single_vs_multiframe_detection}. 
This indicates that the current TBA problem formulation may not be the best alignment of the detection and tracking objectives. 

\begin{table}[t]
\centering
\caption{Comparison of detection performance between the single-frame pretraining and the three-frame fine-tuning. Evaluated on the seven tracking classes on the nuScenes validation set. The best results are typeset in boldface.}
    \begin{tabular}{l|cc}
    \specialrule{1pt}{0pt}{0pt}
    JDT3D variant                          & mAP $\uparrow$ & NDS $\uparrow$ \\
    \hline
    Single-frame                  & \textbf{0.605} & \textbf{0.549} \\
    Three-frame                   & 0.399 & 0.502 \\
    \bottomrule
    \end{tabular}
\label{tab:single_vs_multiframe_detection}
\end{table}

\begin{table}[t]
\centering
\caption{The average confidence of positively matched proposal and track predictions over a three-frame training sequence. There are no track queries in the first frame.}
    \begin{tabular}{c|ccc}
        \specialrule{1pt}{0pt}{0pt}
        Avg. Confidence & Frame 1 & Frame 2 & Frame 3 \\
        \hline
        Positively Matched Proposals & 0.669 & 0.409 & 0.387 \\
        Track Predictions & n/a & 0.759 & 0.760 \\ 
        \bottomrule
    \end{tabular}

\label{tab:avg_pos_scores}
\end{table}

\subsection{Proposal and Track Query Imbalances}

\cref{tab:avg_pos_scores} shows further analysis of the JDT3D outputs. During training, the positively matched proposal queries, or untracked newborn objects, are more confident in the first frame than in subsequent frames. Also, the positively matched track queries, or previously tracked objects, tend to be more confident than the newborn objects.
This indicates that the model has a strong bias toward previously tracked objects. This bias could lead to more missed objects and fewer initialized tracks, explaining the high number of false negatives indicated in \cref{tab:tbd_jdt_nus_test_comparison}.

One hypothesis for this behaviour is that this is caused by the imbalance in training samples assigned to each type of query. \cref{tab:query_assignment_imbalance} shows the average distribution of true positive predictions of both proposal and track queries in the second and third frames of training. Note that the proportion of positively matched proposal queries is significantly lower than for track queries. This is somewhat expected since the average track birthrate per frame is 0.7, so most objects in a scene were present in the previous frame.
This imbalance in positive training samples between the proposal and track queries could explain the confidence imbalance observed.

\begin{table}[t]
    \centering
    \caption{The distribution of positive and negative matches for proposal queries and track queries in the second and third frames of training.}
    \begin{tabular}{l|cc}
    \specialrule{1pt}{0pt}{0pt}
         & PQ Predictions & TQ Predictions \\
         \hline
        Avg Positive Matches & 4 (2\%) & 47 (94\%) \\
        Avg Negative Matches & 196 (98\%) & 3 (6\%) \\
        \bottomrule
    \end{tabular}
    \label{tab:query_assignment_imbalance}
\end{table}

\section{Discussion}
\label{sec:discussion}

Based on our analysis from \cref{tab:jdt3d_tbd_comparison} we show that the reason LiDAR-based TBA methods continue to underperform compared to TBD methods is due to poorer detection performance. In addition, we note that there is a significant drop in the detection performance when fine-tuning the model on multiple frames in \cref{tab:single_vs_multiframe_detection}. In the following sections, we discuss hypotheses to explain the observed results and propose potential improvements to build on JDT3D.

\subsection{Temporal Confusion}
\begin{figure}[tb]
    \centering
    \includegraphics[width=0.75\linewidth]{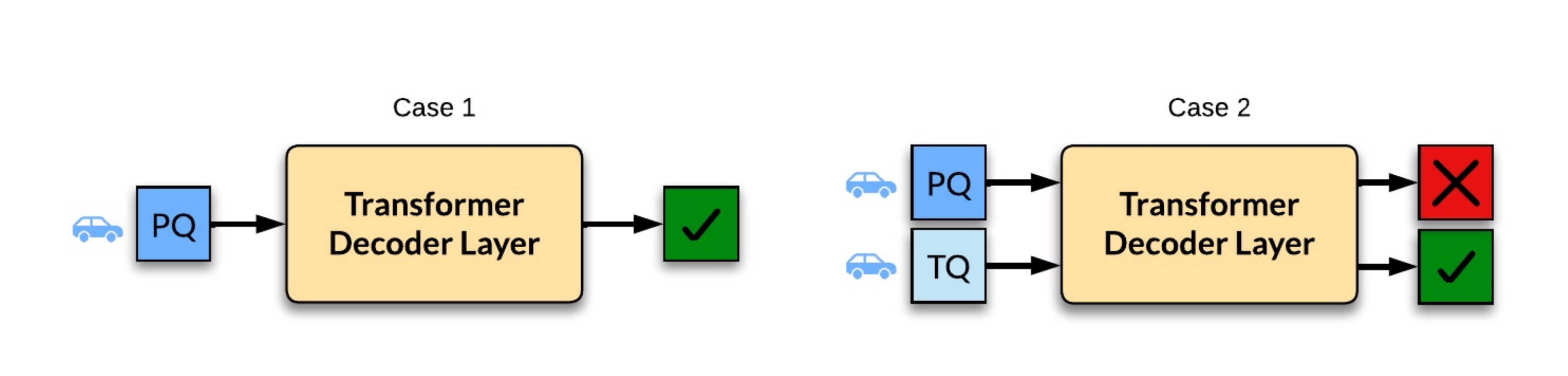}
    \caption{An example of temporal confusion, where the decoder must handle the same proposal query differently based on the presence of track queries. In Case 1, the proposal query should be a positive prediction, while in Case 2, it should be a negative prediction.}
    \label{fig:temporal_confusion}
\end{figure}

Our hypothesis for the observed drop in performance in \cref{tab:single_vs_multiframe_detection} lies in the fact that the problem formulation of TBA is more complex than query-based detection. In query-based detection, the handling of the queries by the decoder is straightforward: if the query represents an object in the scene, it should output a high-confidence bounding box from the query. On the other hand, in TBA, the handling of queries is conditional: depending on whether the decoder receives a track query that represents the same object, the decoder needs to determine whether the incoming proposal query is a duplicate that should have a low confidence or a unique object that should have a high confidencem, as illustrated in \cref{fig:temporal_confusion}.
This complex interaction results in "temporal confusion" such that the expected model behaviour is different depending on the query inputs. 
We hypothesize that the reason temporal confusion has not been an issue in vision-based TBA methods \cite{Pang_2023_CVPR, trackformer_2022_CVPR, Zhang_2022_CVPRW} is due to their larger decoder size enabling more complex reasoning about the queries. To test this hypothesis, further optimizations would be needed to enable a larger decoder with a LiDAR backbone due to hardware constraints.

\subsection{Potential Improvements}
With these results in mind, we propose two potential directions that can be further explored to bridge the gap between JDT and TBD methods.

Firstly, based on the ablation performed to evaluate the training clip size, we observed that three-frame training significantly improved performance over two-frame training, although extending to four frames reduced performance slightly. Further exploration of the effect of clip length on tracking performance may yield insights on how to better leverage longer term temporal information.

Secondly, optimizations to the model size and memory footprint would enable a larger and more complex decoder, that could potentially address the temporal confusion hypothesis.
\section{Conclusion}
\label{sec:conclusion}
This paper presents JDT3D, a novel LiDAR tracking method that makes significant progress in closing the gap between TBD and TBA methods. 
JDT3D is trained using track sampling augmentation, a generalizable data augmentation method that injects extra long-term tracks into training samples. We show that track sampling augmentation significantly improves training performance and convergence, and we hope that future LiDAR-based trackers can leverage these findings. 
As well, we show that the query propagation scheme during training also has a significant impact on the performance, showing that using a confidence threshold to filter low confidence predictions improves tracking performance greatly. Based on our experiments, we hypothesize that the complexity of tracking-by-attention combined with the memory constraints imposed by LiDAR feature extractors limiting the size of transformer decoders is the cause of the drop in detection performance between single and multi-frame variants of JDT3D. We propose further areas of exploration including exploring different training sequence lengths and enabling more decoder layers.

%
%
\bibliographystyle{splncs04}
\bibliography{main}

\newpage
\end{document}